\documentclass[conference]{IEEEtran}
\IEEEoverridecommandlockouts

\usepackage{cite}
\usepackage{amsmath,amssymb,amsfonts}
\usepackage{algorithmic}
\usepackage{graphicx}
\usepackage{textcomp}
\usepackage{xcolor}
\def\BibTeX{{\rm B\kern-.05em{\sc i\kern-.025em b}\kern-.08em
    T\kern-.1667em\lower.7ex\hbox{E}\kern-.125emX}}
\begin{document}

\title{Learning signatures of decision making from many individuals playing the same game\thanks{*Both first authors (mmendelson3@gatech.edu, mazabou@gatech.edu) contributed equally. $\dagger$ Both senior authors (evadyer@gatech.edu, herma686@umn.edu) contributed equally. This work was supported by NIH awards 1R01EB029852-01, R21HM127607, and K23MH050909, NSF award  IIS-2039741, and  the McKnight Foundation.}
}

\makeatletter
\newcommand{\linebreakand}{%
  \end{@IEEEauthorhalign}
  \hfill\mbox{}\par
  \mbox{}\hfill\begin{@IEEEauthorhalign}
}
\makeatother

\author{\IEEEauthorblockN{Michael J. Mendelson$^{1,*}$,
Mehdi Azabou$^{1,*}$, Suma Jacob$^2$, Nicola Grissom$^2$, David Darrow$^2$, \\Becket Ebitz$^3$,
Alexander Herman$^{2,4,\dagger}$, Eva L. Dyer$^{1,\dagger}$}
\vspace{3mm}
\IEEEauthorblockA{
1 - Georgia Institute of Technology, Atlanta, GA, USA\\
2 - University of Minnesota, Minneapolis, MN, USA\\
3 - Universite de Montreal, Montreal, Quebec, CA\\
4 - University of Minnesota School of Medicine, Minneapolis, MN, USA
}}

\maketitle

\begin{abstract}
Human behavior is incredibly complex and the factors that drive decision making—from instinct, to strategy, to biases between individuals—often vary over multiple timescales. In this paper, we design a predictive framework that learns representations to encode an individual's `behavioral style', i.e. long-term behavioral trends, while simultaneously predicting future actions and choices. The model  explicitly separates representations into three latent spaces: the recent past space, the short-term space, and the long-term space where we hope to capture individual differences. To simultaneously extract both global and local variables from complex human behavior, our method combines a multi-scale temporal convolutional network with latent prediction tasks, where we encourage embeddings across the entire sequence, as well as subsets of the sequence, to be mapped to similar points in the latent space. 
We develop and apply our method to a large-scale behavioral dataset from 1,000 humans playing a 3-armed bandit task, and analyze what our model's resulting embeddings reveal about the human decision making process. In addition to predicting future choices, we show that our model can learn rich representations of human behavior over multiple timescales and provide signatures of differences in individuals.
\end{abstract}

\begin{IEEEkeywords}
Human decision making; self-supervised learning; contrastive learning; multi-arm bandit task; behavior analysis
\end{IEEEkeywords}

\section{Introduction}
Human behavior is incredibly complex, with decisions being shaped by a number of factors that can unfold over multiple timescales \cite{pentland1999modeling, silverman2002human}. Building a model that reveals the dynamics of complex behavior can help us discover many new insights into neural function and human decision-making. This has the potential to lead to more effective treatment for neuropsychiatric disorders that affect or impair decision-making.

A core challenge we face in understanding human cognitive processes is modeling behavior in a way that learns a comprehensive representation of an individual's state or `behavioral style'—a broad characterization of how the individual approaches the task over a long timescale—rather than just learning short-term actions. To do this requires a unified view of short-term actions and behavior, as well as long-term trends in behavior and differences between individuals. Recent work in this direction utilizes contrastive learning objectives \cite{task_programming} or reconstruction-based objectives \cite{co2018self,CHEN2021332} to build such representations. However, these methods do not explicitly take into account the multi-timescale nature of behavior and have rarely been applied to human behavior.

In this work, we develop a novel method for extracting the multi-timescale features that drive human behavior. To do so, we design an architecture that explicitly separates behavioral information at  different time scales. We define a self-supervised predictive task for the model, where the network {\em predicts actions at future time points} from three latent representations: one that encodes the {\em recent past}, one that encodes the {\em short-term history}, and one that encodes {\em long-term history} or global information that is consistent throughout a recording session. 

To test our approach, we apply the model to a unique human behavioral dataset, where individuals aim to maximize their reward by choosing among three card decks with randomly drifting binary reward probabilities (Figure~\ref{fig:architecture}). In analyzing our model's performance both at predicting choice at the next timestep and at generating discriminative latent representations across multiple timescales, we demonstrate that our model has a novel ability to learn the multi-timescale features that drive human behavior and decision making.

The contributions and findings of this work include:
\begin{itemize}
    \item A new model to explicitly extract multi-timescale behavioral features in human decision making tasks (Figure \ref{fig:overview}). By bootstrapping across multiple receptive fields we learn explicit latent representations for each timescale.
    \item An interactive visual tool (Figure \ref{fig:overview}) displaying model outputs and dataset metadata. This allows for a novel level of interpretability to our model, a crucial component to analyzing complex human behavior.
    \item When applied to a new human multi-arm bandit dataset, we show how our model can be used to effectively predict future choices and can generate latent embeddings of behavior that have meaningful invariances and reveal signatures of decision making. 
\end{itemize}

\begin{figure*}[h!]
\centering
    \hspace{-4mm}
    \includegraphics[width=\textwidth]{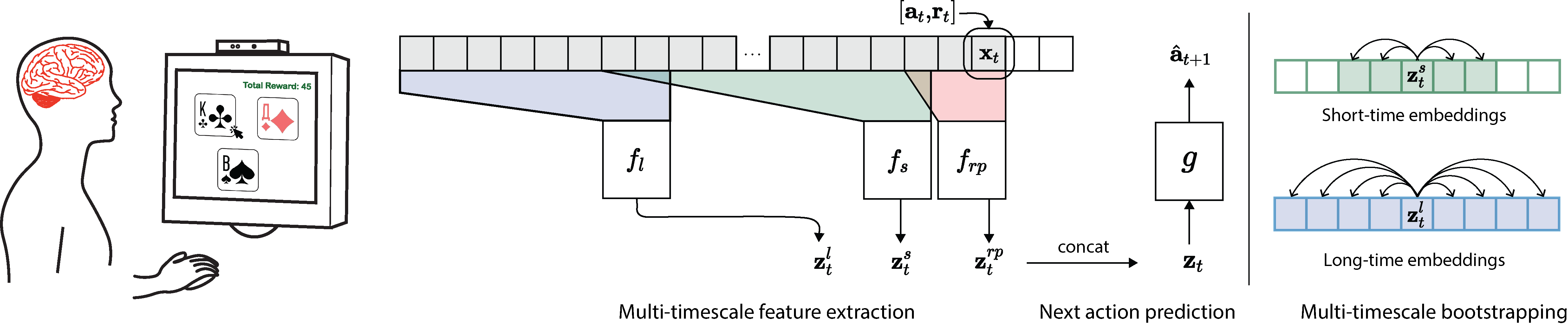}
    \caption{\footnotesize{\em TCN architecture.} (left) We show the 3-armed bandit task. (middle) We depict the multiple latent spaces that are learned by our model and the varying receptive fields (amount of past behavior) each TCN encoder receives. We show the prediction of the next action using classifier g. (right) We combine this architecture with a bootstrapping objective which aims to embed nearby points in time to nearby points in the latent space. We perform bootstrapping across multiple timescales to achieve our objective of forward prediction of actions.
    }
    \label{fig:architecture}
\vspace{-3mm}
\end{figure*}

\section{Methods}

\begin{figure*}[t!]
\centering
    \includegraphics[width=0.98\textwidth]{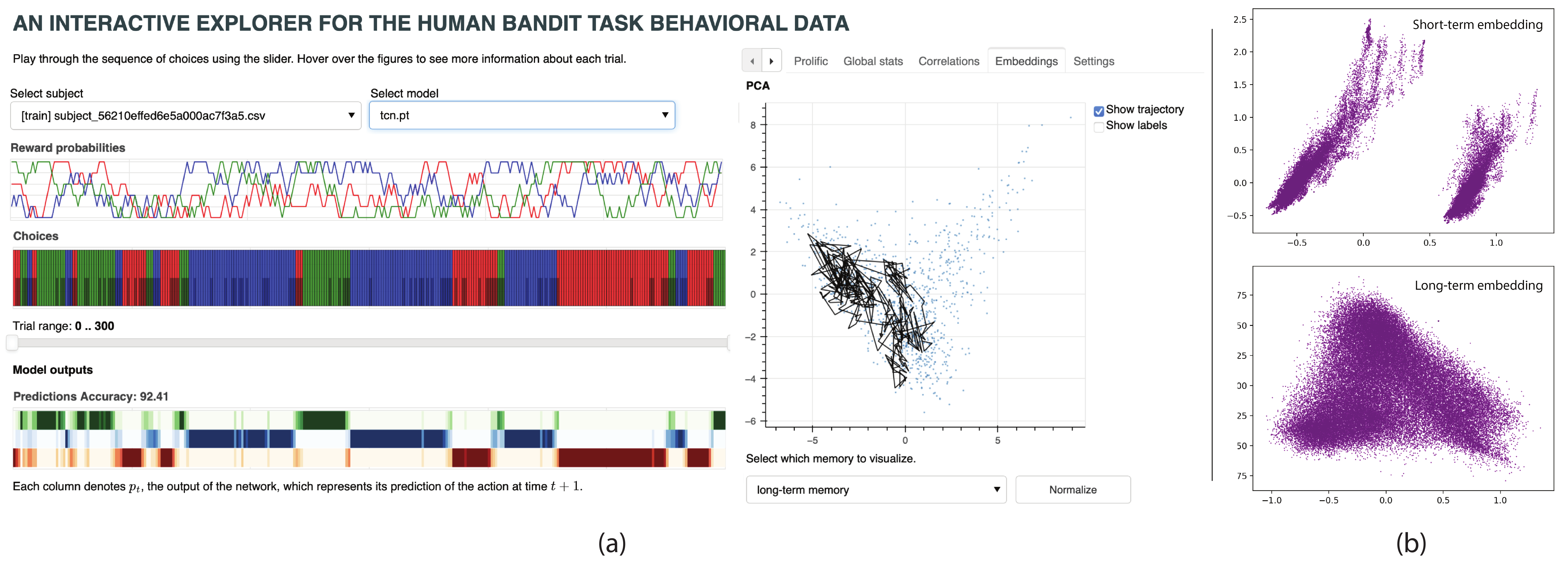}
    \vspace{-3mm}
    \caption{\footnotesize{\em Multi-timescale embedding visualization tool.} (a) We develop a interactive tool for analyzing input data and metadata, as well as output embeddings and predictions. Visualizations can be generated for any subject in the dataset and for any model that has been trained. On the top left of the visualization, you can see the reward probabilities for each arm of the 3-armed bandit task at each time point. Below, you can see the choices made at each time point as well as the models predictions (with the darkness of the color illustrating its confidence). It is possible to zoom in for a more detailed view using the slider. To the right, you can visualize the networks embeddings for any of the encoders, as well as many other statistics. (b) We show the PCA projections of the short-term and long-term embeddings from our model for the entire dataset.
    }
    \label{fig:overview}
\vspace{-2mm}
\end{figure*}

To build a rich space for building predictions from strategy that can arise over different timescales in human behavior, we both build a comprehensive 3-arm bandit dataset and develop a contrastive, self-supervised approach (Figure \ref{fig:architecture}) for human behavior, building on our previous work in multi-task representation learning from animal behavior \cite{azabou2022learning}.

\subsection{Multi-Armed Bandit Dataset}
To study decision making across a large pool of individuals, we collected data from human participants performing a 3-armed bandit task (Figure \ref{fig:architecture}) \cite{vermorel2005multi}. Each arm has an associated Bernoulli distribution parameterized by a stochastically drifting reward probability 
(Figure \ref{fig:overview}), meaning that there is no clear `optimal strategy' and that the player's decisions are purely based on an internal strategy or theory. Each participant iteratively chooses one of the 3 arms—different player card decks—and receives an associated binary reward. After completing an initial demo of 25 rounds, the objective is to maximize the reward collected over the full trial of 300 rounds. The data was collected from 1000 participants (500 male / 500 female) using Prolific, an online platform for recruiting research participants \cite{palan2018prolific}. All participants provided written, informed consent, and all study procedures were approved by the University of Minnesota Institutional Review Board (Study \#00008486). 
Participants were required to reach at least a 42\% reward rate (better than chance) on the training trials, and the probability walk for each participant's 300 trial run was randomly generated to minimize potential biases in response patterns. 

\subsection{Our Approach}

The participant's choices are dictated by their current strategy, which unfolds over a few steps and changes slowly over time (short-term). On the other hand, their behavioral style, which characterizes how they approach the task, manifests through the entire session (long-term). We use this insight to induce different forms of stability and invariance into our learned representations. 

To capture these different factors, we encourage separation across the short-term, long-term, and recent past by coupling (i) a multi-scale architecture with different sized receptive fields into the past, (ii) a latent predictive loss that encourages similarity and invariances across the different latent spaces, and (iii) data-dependent augmentations for further enriching the latents. We now describe all three components.

\subsubsection{Multi-timescale feature extractors}

Let ${\bf a}_t$ be the choice made at timestep $t$, and ${\bf r}_t$ be the reward received. We define an observation as ${\bf x}_t = [{\bf a}_t, {\bf r}_t]$. We aggregate history across different timescales using three encoders with different receptive fields: the recent past encoder observes observations in the $[t-T_{rp}, t]$ range, the short-term encoder observes observations in the $[t-T_s, t-T_{rp}+1]$ range, and the long-term encoder in the $[t-T_l, t-T_s+10]$ range. In our experiments on a human multi-arm bandit dataset, we set $T_{rp}=3$, $T_{s}=20$, and $T_{l}=100$. 

In order to control the receptive field or the range of observable history, we choose to use the temporal convolutional network (TCN) \cite{bai2018tcn} as our building block. 
The three encoders respectively produce embeddings ${\bf z}_{t}^{rp}$, ${\bf z}_{t}^{s}$ and ${\bf z}_{t}^{l}$. All three of these feature embeddings are concatenated to produce one multi-timescale feature embedding:
\begin{equation}
    {\bf z}_t = \mathbf{concat}[{\bf z}_{t}^{rp}, {\bf z}_{t}^{s}, {\bf z}_{t}^{l}]
\end{equation}
A detailed diagram of the archetecture can be seen in Figure~\ref{fig:architecture}.

\subsubsection{Multi-timescale latent predictive loss}
To train our model, we propose a multi-scale latent prediction objective \cite{grill2020bootstrap, brave} that encourages similarity between feature embeddings at different time horizons. For the short-term embedding, we select two timesteps, $t$ and $t+\Delta_s$, that are within a window $\Delta_s$ of each other, and push their short-term embeddings closer to each other. 
This window, $[-\Delta_s, \Delta_s]$, defines the positivity range over which samples should have similar short-term embeddings. 
We do the same for the long-term embedding, but now consider a much longer window. In this case, we will encourage the long-term representation to be similar over each individual's behavioral sequence and thus build a sequence-level space for comparing different individuals.

For the short- and long-term embedding, we use respective predictors $q_s$ and $q_l$ that take in embeddings ${\bf z}^s_{t}$ and ${\bf z}^l_{t}$ and learn to predict the latent representation of a nearby point in time. The window over which we decide to group points as neighbors depends on the timescale over which we want to encourage smoothness. 

Let $\Delta_s$ denote the window over which we encourage similarity for the short-term embeddings. We can write out the loss for our short-term embedding as follows: 
\begin{equation}
    \mathcal{L}_{r}^s = \left \|\frac{q_s({\bf z}_{t}^{s})}{\|q_s({\bf z}_{t}^s)\|_2} - \mathrm{sg}\left [ \frac{{\bf z}_{t+\Delta_s}}{\|{\bf z}_{t+\Delta_s}\|_2} \right] \right \|_2^2
\end{equation}
with $\mathrm{sg}[\cdot]$ denoting the stop gradient operator. Similarly, we define the long-term loss as $ \mathcal{L}_{r}^l$ and increase the size of $\Delta_l$ to a much longer window size to encourage a more global stability in the long-term space.

Next, we combine our short- and long-term embeddings with a small recent past embedding to predict future actions and also mask the immediate past choice from observation in the short- and long-term spaces, making their latent invariant to the behavior at the most recent timesteps. Given observations ${\bf x}_t$ of the behavior at $t=0$ through $t$, the model extracts a feature embedding ${\bf z}_t$, and a classifier $g$ predicts the next action ${\bf a}_{t+1}$. 
The action prediction loss is thus written as follows:
\begin{equation}
    \mathcal{L}_{p} = \mathrm{cross\ entropy}(g({\bf z}_t), {\bf a}_{t+1})
\end{equation}
In learning how to predict future actions, the network learns how to model behavioral dynamics.

The total loss can be written as:
\begin{equation}
\mathcal{L} = \mathcal{L}_p + \alpha (\mathcal{L}_r^l + \mathcal{L}_r^s),
\end{equation}
where $\alpha$ is a hyperparameter that weights the emphasis on the latent vs. forward prediction terms.

\subsubsection{Augmentations and permutation invariance}
In addition to training on the original sequence of choices from an individual, we also train the model on every possible permutation of the choice identifiers across each sequence under the assumption that the specific identity of the choice, as long as it is consistent throughout entire sequences, is not relevant to the strategy. This increases the amount of training data by 6 fold. As our goal is to learn a representation of each individual's behavioral style, the specific identity of their choice is not relevant, and the model should be invariant to it.

\section{Results}
To evaluate the efficacy of our model at learning rich representations of behavior, we perform several experiments to evaluate the model's different latent spaces, as well as examine our dataset and the model's response in an interactive visualization tool (Figure \ref{fig:overview}).

\subsection{Experimental setup}
We split our dataset into a train and test set (80/20) across individuals, and report the future choice prediction accuracy on the held-out test set. We compare to a multi-layer perceptron (MLP) baseline with the same prediction task, as well as perform multiple ablations to our model, including a single TCN encoder. Each of the models is trained over 1000 epochs using an AdamW optimizer \cite{loshchilov2017decoupled} with a learning rate of 0.01. To demonstrate the separability of choices across the model's different latent spaces (Table~\ref{results-table-2}), we fit a simple linear layer on top of the model's embeddings to classify the individual's choice. To quantify the separability of clusters in the different embedding spaces, we also compute the mean silhouette coefficient of each respective embedding space, after being normalized and reduced to 5D using PCA. In the case of a 3-armed bandit task, random prediction chance is 1/3.

\subsection{Interactive visualization tool}
We developed an interactive visualization tool (Figure \ref{fig:overview}) to help inform our approaches and help us interpret our results. Due to the complex nature of human behavior, interpretability is crucial when modeling it. We do this by visualizing informative metadata and latent representations of our model. This provides us with a more holistic way of evaluating our model than through traditional evaluation metrics alone.

\begin{table}[!ht]
\begin{center} 
\vspace{-2mm}
\caption{Choice prediction accuracy for different models.} 
\label{results-table} 
\begin{tabular}{ll} 
\hline
Model      &  Accuracy   \\
\hline
Our model  &   76  \\
MLP Baseline        &   51  \\
Our model - no contrastive &   75 \\
Our model - no permutation augmentation  &   73  \\
Our model - recent-past encoder only &   72  \\
\hline
\end{tabular} 
\end{center} 
\end{table}
\vspace{-4mm}

\begin{table}[!ht]
\begin{center} 
\caption{Separability of choices in different parts of the latent space.} 
\label{results-table-2} 

\begin{tabular}{lll} 
\hline
Embedding Space     &  Acc  & Score \\
\hline
Full Space        &   63  & 0.32\\
Long-term + Short-term embeddings   &   58  & -0.02\\
Long-term embedding   &   52  & -0.01\\
Short-term embedding   &   54  & -0.03\\
Recent past embedding    &   51  &  0.51\\
\hline
\end{tabular} 
\end{center} 
\end{table}

\subsection{Analysis of ablated representations}
In our analysis, we demonstrate the performance for different ablations of our model as accuracy scores (Table~\ref{results-table}). Illustrating the performance of different ablations of the model can help us interpret the effectiveness of individual components of our model, and help inform how well the model learns the individual's behavior, both short and long term. Notably, as the pretext task is predicting the individual's choice (i.e. predicting at every time step), it is likely that a high accuracy is more indicative of rich short-term behavioral features than long-term.

In Table~\ref{results-table} we see that the TCN with permutation invariance and contrastive loss performs the best. This is expected as the contrastive loss encourages rich representations and discriminative features, thus a more complete representation of both an individual's behavioral style and short-term behaviors that inform their next choice. The contrastive loss also allows for greater seperability between timescales. Similarly, the TCN with permutation invariance but no contrastive loss performs comparably. Permutation invariance encourages rich representations, as it discourages the network from learning class-specific choice patterns but rather encourages learning of behavioral patterns. Even the ablated models without permuatation invariance and with only the recent-past encoder significantly outperform our baseline.

\subsection{Choice separability}
In addition to our model's classification performance, we demonstrate the separability of choice in our model's different latent spaces as the accuracy when the model's features are fed through a simple linear classifier (Table~\ref{results-table-2}). This helps us interpret the model's learned representation over each explicitly defined behavioral timescale.

Table~\ref{results-table-2} shows that using the full space performs highest for classifying choice, as expected. However, the recent past performs the lowest, indicating that the model learns representations of the individual's behavioral style, and is not simply predicting based on their most recent choices. This is further confirmed by the fact that the model performs comparably to the full space with just the recent past removed.

\subsection{Clustering the latent spaces}
Finally, we report the silhouette score for different latent spaces (Table~\ref{results-table-2}), which demonstrates how well each space is clustered with respect to choice, or each space's efficacy at classifying an individual's choice at the next time step (which is not necessarily indicative of the richness of each latent space).

In Table~\ref{results-table-2} we see that the recent past embedding performs the highest followed by the full space. These results are expected as choice at the next timestep is likely most dependent on recent choices. In turn, it is likely invariant to the choice, say, 50 timesteps ago. Hence, the lower scores of the short-term and long-term embeddings indicates that the longer timescale embedding spaces are learning other information about the individual's behavior, that is not directly related to choice—perhaps the individual's behavioral style. We plan to further quantify this observation in future work.

\section{Discussion}
In this work, we introduced a new model for learning separable multi-scale representations of behavior.
From our results, we can see that our model is able to learn rich representations of complex human behavior over multiple timescales simultaneously. This provides novel information about individual's immediate actions and long-term behavioral style, providing a broad characterization of an individual's behavior and decision-making, as well as population-wide behavioral characteristics.

Variability of human behavior is likely to be driven by a number of complex factors that unfold over different timescales. Thus, having ways to model behavior and discover differences in behavioral repertoires, actions, and `styles' could provide invaluable insights into differences in an individual's cognitive state, and even detect signatures of cognitive impairment, especially at large scales and over heterogeneous populations. This could reveal insights about human behavior and decision-making that cannot be revealed through more traditional approaches or without the power of deep learning.

\bibliographystyle{./IEEEtran}
\bibliography{./IEEEabrv, ./main_bib}

\end{document}